\documentclass[submission,copyright,creativecommons]{eptcs}
\providecommand{\event}{Doctoral Consortium of the International Conference on Logic Programming 2022} 

\usepackage{pgfplots}
\pgfplotsset{compat=1.16}
\pgfplotstableread{
1	3	3	7	3
2	5	3	7	3
3	6	2	8	3
4	7	3	12	3
5	4	2	5	3
6	5	2	3	3
7	4	3	4	3
8	5	3	4	3
9	8	3	1	3
10	6	2	1	3
11	3	3	3	3
12	2	3	2	3
13	2	2	3	3
14	3	3	5	3
15	2	3	2	3
16	1	3	1	3
17	1	3	3	3
18	2	3	0	3
19	1	3	2	3
20	0	3	0	3
21	0	3	1	3
22	0	3	3	3
23	0	2	0	3
24	1	3	0	3
25	0	3	0	3
26	0	2	1	3
}\dataset

\title{Planning and Scheduling in Digital Health with Answer Set Programming}
\author{Marco Mochi
\institute{University of Genoa\\ Genova, Italy}
\email{S4094982@studenti.unige.it}}
\def\titlerunning{Planning and Scheduling in Digital Health with Answer Set Programming}
\def\authorrunning{Marco Mochi}

\begin{document}
\maketitle

\begin{abstract}
In the hospital world there are several complex combinatory problems, and solving these problems is important to increase the degree of patients' satisfaction and the quality of care offered.
The problems in the healthcare are complex since to solve them several constraints and different type of resources should be taken into account. Moreover, the solutions must be evaluated in a small amount of time to ensure the usability in real scenarios.
We plan to propose solutions to these kind of problems both expanding already tested solutions and by modelling solutions for new problems, taking into account the literature and by using real data when available.
Solving these kind of problems is important but, since the European Commission established with the General Data Protection Regulation that each person has the right to ask for explanation of the decision taken by an AI, without developing Explainability methodologies the usage of AI based solvers e.g. those based on Answer Set programming will be limited. Thus, another part of the research will be devoted to study and propose new methodologies for explaining the solutions obtained.  
\end{abstract}

\section{Introduction and problem description}

The usage of Answer Set Programming has proved to be a viable solutions to many healthcare problems \cite{DBLP:conf/aiia/AlvianoBCDGKMMM20} thanks to the availability of efficient solvers and the rich syntax. In this section two healthcare problems are presented: The Chemotherapy Treatment Scheduling problem and the Operating Rooms Scheduling (ORS) problem. These two problems have characteristics and complexity such that can be used as examples of healthcare problems. After presenting these two problems we introduce the problem of explainability. 

The Chemotherapy Treatment Scheduling (CTS) \cite{hahn-goldberg_dynamic_2014,huang_alternative_2017,huggins_2014,sevinc_algorithms_2013} problem consists of computing a schedule for patients requiring chemotherapy treatments.
The CTS problem is a complex problem for oncology clinics since it involves multiple resources and aspects, including the availability of nurses, chairs, and drugs.
Each patient and treatment have different priorities and the scheduler must take care of the differences to produce a proper solution. Producing an optimal solution is crucial for improving the usage of the resources and the degree of satisfaction of patients and nurses.
Various studies, also in the context of the COVID19 emergency \cite{kumar_treatment_2020,sud_collateral_2020}, have shown how delays in cancer surgeries and treatments have a significant adverse impact on patient survival. Moreover, the impact of the delays is greater for some type of cancer, thus the importance of implementing different priorities. 
Many works focused on the CTS problem but, as noted in \cite{Lame_2016}, the majority of these works proposed a solution focusing on the starting time of the treatments ignoring the earlier phases, while other work ignored the demand of bed or chair, so what lacks in the context of the CTS problem is a solution that involves the resources needed by the patients and the phases before the treatment.
The motivation to focus on the phases before the treatment is that, in many hospitals, multiple departments share the same resources.
In particular, we identify four phases for each patient:
1) the registration to the Hospital reception;
2) a blood collection;
3) a medical check; and
4) the therapy.
Other than satisfy all the constraints of the problem, to obtain an optimal solution the scheduler has to minimize the number of patients assigned to the wrong resources and the number of concurrent patients starting phase 2. 
In our paper \cite{dodaro_galata_grioni_maratea_mochi_porro_2021}, differently from other works, we defined the requirements according to the indications of the S. Martino Hospital in Genova, Italy. Moreover, we used real data for the empirical analysis.

Most modern hospitals are also characterized by a very long surgical waiting list. The elements of the waiting list are called registrations. Each registration links a particular surgical procedure, with a predicted duration, to a patient.
The overall goal of the ORS problem is to assign the maximum number of registrations to the operating rooms (ORs). As a first requirement, the assignments must guarantee that the sum of the predicted duration of surgeries assigned to a particular OR shift does not exceed the length of the shift itself.

Moreover, registrations are not all equal: they can be related to different pathologies and they can be in the waiting list for different periods of time. These two factors can be unified in a single concept: priority. The registrations with the highest priority gathers either very urgent registrations or the ones that have been in the waiting list for a long period of time; it's required that these registrations are all assigned to an OR. Then, the registrations with lower priorities are assigned according to the ORs capacity.
Each operating room is assigned to a particular specialty, so for each registration, requiring a specific surgery and thus a specific specialty, the solution must assign the patient to the operating room assigned to the same specialty requested by the patient.

After having presented two healthcare related problems, it's important to present another path of the research that will be conducted: Explainability.
Explainable Artificial Intelligence (XAI) is a field of AI that allows human-understandable solutions. In the healthcare domain, proposing a black-box model as solution will not be enough in the future, both because the patients and the operators need to know how and why a certain solution and a certain decision was made and because the European Commission has established, with the General Data Protection Regulation, that each person has the right to ask for explanation of the decision taken by an AI. Moreover, even the USA Department of Defense is going to the same direction.
Without developing Explainability methodologies the usage of AI based solvers will be limited, thus, both for ethical and legal reason the implementation of explainability techniques will be crucial in all the fields in which AI could be applied and even with more urgency in the healthcare domain.

\section{Overview of the existing literature}

\subsection{Healthcare domain problems}
In this section are presented the state of the art of the two problems considered before.
\cite{sevinc_algorithms_2013} addressed the CTS problem through a two-phase approach. In the first one an adaptive negative-feedback scheduling algorithm is adopted to control the load on the system, while in the second phase two heuristics based on the ‘Multiple Knapsack Problem’ have been evaluated to assign patients to specific infusion seats. The overall design has been tested at a local chemotherapy center and has yielded good results for patient waiting times, orderly execution of chemotherapy regimen and utilization of infusion chairs. \cite{huang_alternative_2017} developed and implemented a model to optimize safety and efficiency in terms of
staffing resource violations measured by nurse-to-patient ratios throughout the workday and at key points during treatment to decide when to schedule patients according to their visit duration. The optimization model was built using Excel Solver.
\cite{hahn-goldberg_dynamic_2014} addressed in particular dynamic uncertainty that arises from requests for appointments that arrive in real time and uncertainty due to last minute scheduling changes through a proactive template of an expected day in the chemotherapy centre using a deterministic optimization model updated, to accommodate last minute additions and cancellations to the schedule, by a shuffling algorithm. \cite{huggins_2014} presented a mixed-integer programming optimization model developed with the objective of maximizing resource utilization, while balancing human workload, in particular taking into account variability in length of treatment, increased patient demand, and resource limitations.
Compared to the papers mentioned above, our solution tackles further aspects.
For the scheduling problem, in addition to taking into account the starting time of treatments, we also manage the previous phases (blood collection and visit). 
We also minimize the number of patients who request blood collection at the same time.
Considering these previous phases is important because, as mentioned in \cite{Lame_2016}, usually the CTS problem involves many departments and knowing when a resource will be used is vital for efficient planning.

For the ORS problem, \cite{aringhieri_two_2015} addressed the scheduling problem, described as the allocation of OR time blocks to specialties together with the subsets of patients to be scheduled within each time block over a one week planning horizon. They developed a 0-1 linear programming formulation of the problem and used a two level meta-heuristic to solve it. Its effectiveness was demonstrated through extensive numerical experiments carried out on a large set of instances based on real data. In \cite{landa_hybrid_2016}, the same authors introduced a hybrid two-phase optimization algorithm which exploits the potentiality of neighborhood search techniques combined with Monte Carlo simulation, in order to solve the joint advance and allocation scheduling problem taking into account the inherent uncertainty of surgery durations. \cite{abedini_operating_2016} developed a bin packing model with a multi-step approach and a priority-type-duration(PTD) rule. The model maximizes utilization and minimizes the idle time, which consequently affects the cost at the planning phase and was programmed using MATLAB. \cite{molina-pariente_new_2015} tackled the problem of assigning an intervention date and an operating room to a set of surgeries on the waiting list, minimizing access time for patients with diverse clinical priority values. The algorithms used to allocate surgeries were various bin packing (BP) operators. They adapted existing heuristics to the problem and compared them to their own heuristics using a test bed based on the literature. The tests were performed with the software Gurobi.

\subsection{Explainability}

In the last few years the interest on Explainable Artificial Intelligence (XAI) \cite{BARREDOARRIETA202082, doi:10.1126/scirobotics.aay7120} has increased as a result of the increased interest on Artificial Intelligence based solutions. Nevertheless, many works still highlight the lack of a clear definition of what does it mean to "explain" a model and tried to define it \cite{DBLP:journals/corr/Miller17a,DBLP:journals/corr/abs-1811-01439}. What can be seen as a goal of explainability techniques is to explain why a certain event occurred or why a certain event is not in the solution. Different kind of strategies were adopted to answer these questions. In this section we will cite some works about explainability in logic programming. For example, some authors focused their work on "Contrastive explanations" \cite{DBLP:journals/corr/Miller17a} or "Justification" \cite{DBLP:journals/corr/CabalarFF14}. In particular, in the works focused on "Contrastive explanations" the explainee is interest on knowing why a certain event occurred instead of another one or on knowing why with the same model and with two similar input the occurred events were different, while, in the works that deal with Justification, the model explains an event by observing all the facts or events that caused the event to be true.
Moreover, in complex problems involving several constraints, it is not always easy to understand why a certain problem is unsatisfiable or why a certain decision was made. So, to allow the users to use the models in the best condition the AI should provide some sort of explanations of the decision taken and provide debugging tools \cite{DBLP:conf/aaai/GebserPST08,DBLP:journals/corr/abs-1808-00417}. 

\section{Goal of the research}

\subsection{Healthcare domain problems}

One of the goals is to propose solutions to healthcare problems using ASP.
Some problems could be extensions of already presented problems such as the Operating Room Scheduling. For example, starting from works such as \cite{DBLP:conf/aiia/DodaroGMP18}, we extended it to enlarge the scope of the solver by including the management of the beds in special care units, for example, patients that have had complicated surgery would need a bed in specific areas, and the solver should assign the day of surgery considering also the bed available in these areas. Other special care units could be added too, e.g care units that are needed by infants or particularly ill patients. For this problem we want to define a model that could fit the constraints and optimization of many special care units. To develop this model we should firstly define all the requirements of the special care units, then, having defined the requirements we should generalize the constraints in common between the special care units and manage the more specific constraints required by the different care units. Moreover, we plan to enlarge the model for the CTS problem too. In particular, for the CTS problem could be considered other constraints that weren't considered in the previous work. A new constraints could be defined considering the different requirements of the drugs needed by the patients, that could be available to the hospital just after the morning, or by the condition of the patients, that could require that a patient is assigned in a room with a scalp cooling or need to be isolated from the other patients. 

Other problems are entirely new problem, such as the Pre-Operative Assessment Clinic scheduling problem, that is the problem of assigning patients to a day in which they will do all the exams required before a surgery. This is a complex problem since the scheduler has to schedule the patients to a valid day not only considering the due date of each patient but also considering that each exam will require the activation of an exam area and thus the scheduling of doctors to the area.

\subsection{Explainability}

Starting from the work \cite{DBLP:journals/corr/abs-1808-00417}, in which was presented a debugger for non-grounded ASP programs, and the work \cite{DBLP:conf/aiia/BertolucciDGMPR21}, that explained why certain atoms weren't in the solution of a healthcare problem, we want to enlarge and further develop the techniques used in these two papers. 

In particular, in \cite{DBLP:journals/corr/abs-1808-00417} was proposed a solution that was able to identify errors and bug in a model through consecutive queries regarding the expected Answer Set. The solution proposed in the paper is able to identify bug and errors when there are no Answer sets, identifying the rules and the atoms causing it. To identify the bugs the solution proposed looking for the minimal reason that caused the unsatisfiability and highlight it to the user. Then, through consecutive queries the model get the background knowledge by the user and, by implementing this additional information, the model looks again for the minimal reason that caused the unsatisfiability.
Our goal is to use the same approach used in the paper and try to generalize it in order to identify why a certain atom is in an Answer Set. Moreover, in the actual solution choice rules are not supported while we plan to support it in our solution. 

In the same way, in \cite{DBLP:conf/aiia/BertolucciDGMPR21}, an explanability layer was added to a solver for the ORs scheduling problem, that explain to the users why the scheduler was not able to reach a valid solution. In particular, when the solver finds a valid solution, this solution is presented to the users while, when the solver found no solutions the explainability framework is called and try to find the sets of atoms that caused the problem to be unsatisfiable. This solution is very helpful for the users since, instead of just knowing that the solver wasn't able to find a solution, thanks to the explanation, they could have hints on how change the instances to solve the problem. Nevertheless, such as the work presented before, this solution is not able to explain and justify why a certain atom is in the solution, indeed the explainability framework is called just when the solver finds no solutions. 

So, the goal for the works in explainability is to develop and propose methodologies, using techniques inspired by the two papers presented before, to explain not only why the solver couldn't find a solution, but also explain why a certain atom is in the solution and then why a certain atom instead of another one.

\section{Current status of the research and preliminary results}

\subsection{Healthcare domain problems}

For the healthcare problems we have already published some works. In particular, one of the main work that we published is \cite{dodaro_galata_grioni_maratea_mochi_porro_2021}, that enriched a previous work \cite{DodaroGMMP20}. The paper was presented at ICLP 2021 and was awarded as the best application paper. 

The first optimization criteria in the CTS problem is to assign the preferred resource (bed or chair) to as many patients as possible. Our solution is able to assign all the patients to the required resource while, in the results obtained by the S. Martino Hospital they used some \emph{virtual chairs}, meaning that they weren't able to assign the patients to a real resource since the patients weren't optimally distributed.

\begin{figure}[t!]
\begin{tikzpicture}[scale=0.58]
\begin{axis}[ybar,
    width=1\textwidth,
	height=0.7\textwidth,
	font=\large,
	x = 0.4cm,
	enlarge x limits=0.02,
    ymin=0,
    ymax=10,        
    ylabel={Patients},
    major tick length=1pt, 
	xtick=data,
    xticklabels = {
        07:40, 07:50, 08:00, 08:10, 08:20, 08:30, 08:40, 08:50, 09:00, 09:10, 09:20, 09:30, 09:40, 09:50, 10:00, 10:10, 10:20, 10:30, 10:40, 10:50, 11:00, 11:10, 11:20, 11:30, 11:40, 11:50
    },
    x tick label style={rotate=90},
    bar width=0.1cm,
    title={(a) Best performance}
]
\addplot[draw=black,fill=blue] table[x index=0,y index=3] \dataset;
\addlegendentry{S. Martino scheduling}
\addplot[draw=black,fill=red] table[x index=0,y index=4] \dataset;
\addlegendentry{ASP scheduling}
\end{axis}
 \end{tikzpicture}
\begin{tikzpicture}[scale=0.58]
\begin{axis}[ybar,
    font=\large,
    width=1\textwidth,
	height=0.7\textwidth,
	x = 0.4cm,
	enlarge x limits=0.02,
    ymin=0,
    ymax=10,        
    ylabel={},
    major tick length=1pt, 
	xtick=data,
    xticklabels = {
        07:40, 07:50, 08:00, 08:10, 08:20, 08:30, 08:40, 08:50, 09:00, 09:10, 09:20, 09:30, 09:40, 09:50, 10:00, 10:10, 10:20, 10:30, 10:40, 10:50, 11:00, 11:10, 11:20, 11:30, 11:40, 11:50
    },
    x tick label style={rotate=90},
    bar width=0.1cm,
    title={(b) Worst performance}
]
\addplot[draw=black,fill=blue] table[x index=0,y index=1] \dataset;
\addlegendentry{S. Martino scheduling}
\addplot[draw=black,fill=red] table[x index=0,y index=2] \dataset;
\addlegendentry{ASP scheduling}
\end{axis}
 \end{tikzpicture}
\caption{\label{fig:compare} Number of patients assigned to their phase 2 for each time slot in a day, comparing our scheduling with the one of the S. Martino Hospital. Figure (a) [resp. (b)]   compares our best [resp. worst] result with the one of the Hospital on the same day.}
\end{figure}
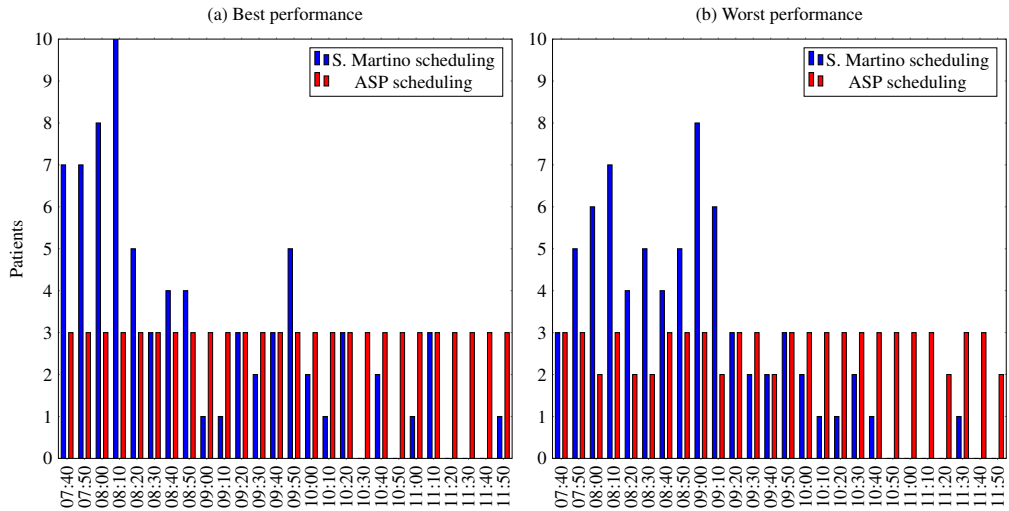

The second optimization criteria is to have a more equally distributed affluence of patients during their phase 2 for each time slot.
We compared our results with the schedule of the Hospital. The results are presented in Figure~\ref{fig:compare}, which details the number of patients starting phase 2 in each time slot; the figure compares our best (a) and worst (b) results with those of the Hospital in the same days. In particular, the figure shows that our solution was able to schedule the patients equally through the day while, the solution of the Hospital led to imbalances and peak of concurrent patients.
The difference between the solution of the automated scheduler and the schedule produced by the Hospital is still great even in the worst results.

Moreover, two preliminary versions of a solution of the Operating Room scheduling problem with care unit management \cite{DBLP:conf/aiia/GalataMMMP21} and of the Pre-Operative Clinic Assessment scheduling problem \cite{DBLP:conf/aiia/CarusoGMMP21} have been published and presented at the 28th RCRA International workshop and at the 9th Italian Workshops on Planning and Scheduling, respectively. The first results obtained in \cite{DBLP:conf/aiia/GalataMMMP21} showed that the solution proposed was able to reach the same solution obtained by the model without the management of the beds in the special care units, thus the solution was promising and we will start from this model to develop a more generalized model.

\subsection{Explainability}

For the explainability problem, as said in the previous section, we want to start from two already published papers and develop new methodologies to address the problem of explainability through justification. In particular, we want to enlarge the work \cite{DBLP:journals/corr/abs-1808-00417} by allowing the usage of the debugger even when the solver finds a solution. To reach this goal, we are working on a model that gets as input an atom or more atoms. To justify the truth value of the atoms we plan to extend the original encoder with the negated value of the atoms required by the users and the facts that led to these atoms. Finally, by finding the minimal reason of unsatisfaiability of this new problem, the model should iterate the process considering the new found atoms that caused the unsatisfiability until all the required atoms and the found ones are justified by some facts. 

\section{Bibliography}
\bibliographystyle{eptcs}
\bibliography{main}
\end{document}